\documentclass{article}
\usepackage{ijcai23}

\usepackage{times}
\usepackage{soul}
\usepackage{url}
\usepackage[hidelinks]{hyperref}
\usepackage[utf8]{inputenc}
\usepackage[small]{caption}
\usepackage{graphicx}
\usepackage{amsmath}
\usepackage{amsthm}
\usepackage{booktabs}
\usepackage[switch]{lineno}

\usepackage{amssymb}
\usepackage{mathtools}
\usepackage[ruled,linesnumbered,noend]{algorithm2e}
\usepackage{subfig}

\newcommand\cmtsize[1]{{\footnotesize{#1}}}
\DeclareMathOperator*{\argmin}{argmin}

\DeclareMathOperator{\MLP}{MLP}
\DeclareMathOperator{\SA}{SA}

\DeclareMathOperator{\MEAN}{MEAN}
\DeclareMathOperator{\MAX}{MAX}
\DeclareMathOperator{\IR}{\mathbb{R}}
\DeclareMathOperator{\G}{\mathcal{G}}

\newcommand\blfootnote[1]{%
	\begingroup
	\renewcommand\thefootnote{}\footnote{#1}%
	\addtocounter{footnote}{-1}%
	\endgroup
}

\title{
	Neural Capacitated Clustering
}

\author{
Jonas K. Falkner
\And
Lars Schmidt-Thieme
\affiliations
Institute of Computer Science, University of Hildesheim, Hildesheim, Germany
\emails
\{falkner, schmidt-thieme\}@ismll.uni-hildesheim.de
}

\begin{document}

\maketitle

\begin{abstract}
Recent work on deep clustering has found new
promising methods also for constrained clustering
problems. Their typically pairwise constraints often can be used to guide
the partitioning of the data. Many problems however,
feature cluster-level constraints, e.g.\ the Capacitated
Clustering Problem (CCP), where each point has a
weight and the total weight sum of all points in each cluster
is bounded by a prescribed capacity.
In this paper we propose a new method for the CCP,
\textit{Neural Capacited Clustering}, that learns a neural network
to predict the assignment probabilities of points to cluster
centers from a data set of optimal or near optimal past
solutions of other problem instances.
During inference, the resulting scores are then used in an
iterative k-means like procedure to refine the assignment
under capacity constraints.
In our experiments on artificial data and two
real world datasets our approach outperforms several
state-of-the-art mathematical and heuristic solvers from
the literature.
Moreover, we apply our method in the context of a
cluster-first-route-second approach to the 
Capacitated Vehicle Routing Problem (CVRP)
and show competitive results on the well-known
Uchoa benchmark.    
\end{abstract}

\section{Introduction}
\blfootnote{Paper accepted at IJCAI 2023:\\ \ \ {\url{https://ijcai-23.org/main-track-accepted-papers/}}}
In recent years much progress has been achieved in applying deep learning methods to solve classical clustering problems. 
Due to its ability to leverage prior knowledge and information to guide the partitioning of the data, constrained clustering in particular has recently gained increasing traction. It is often used to incorporate existing domain knowledge in the form of pairwise constraints expressed in terms of \textit{must-link} and \textit{cannot-link} relations~\cite{wagstaff2001constrained}.
However, another type of constraints has been largely ignored so far: cluster level constraints. This type of constraint can for example restrict each assignment group in terms of the total sum of weights which are associated with its members. The simplest case of such a constraint is the maximum size of the cluster, where each point exhibits a weight of one. In the more general case, weights and cluster capacities are real valued and can model a plenitude of practical applications. 

\noindent
Machine learning approaches are particularly well suited for cases in which many similar problems, i.e.\ problems from the same distribution, have to be solved.
In general, most capacitated mobile facility location problems (CMFLP)~\cite{raghavan2019capacitated} represent this setting when treating every relocation as a new problem. This is e.g.\ the case when considering to plan the location of disaster relief services or mobile vaccination centers 
for several days, where the relocation cost can be considered to be zero since the team has to return to restock at the end of the day.
Other applications are for example the planning of the layout of factory floors which change for different projects or logistics problems like staff collection and dispatching~\cite{negreiros3capacitated}. Thus, we can utilize supervised learning to learn from existing data how to solve new unseen instances.

The corresponding formulation gives rise to well-known problems from combinatorial optimization, the \textit{capacitated $p$-median problem} (CPMP)~\cite{ross1977modeling} where each center has to be an existing point of the data and the \textit{capacitated centered clustering problem} (CCCP)~\cite{negreiros2006capacitated} where cluster centers correspond to the geometric center of their members.
The general objective is to select a number $K$ of cluster centers and find an assignment of points such that the total distance between the points and their corresponding centers is minimized while respecting the cluster capacity. Both problems are known to be \textit{NP}-hard and have been extensively studied~\cite{negreiros2006capacitated}.

\subsubsection{Contributions}
\begin{itemize}
	\item We propose the first approach to solve general Capacitated Clustering Problems based on deep learning.
	\item Our problem formulation includes well-known problem variants like the CPMP and CCCP as well as more simple constraints on the cluster size.
	\item The presented approach achieves competitive performance on several artificial and real world datasets, compared to methods based on mathematical solvers while reducing run time by up to one order of magnitude.
	\item We present a cluster-first-route-second extension of our method as effective construction heuristic  for the CVRP.
\end{itemize}


\section{Background}
A typical clustering task is concerned with the grouping of elements in the given data and is normally done in an unsupervised fashion. This grouping can be achieved in different ways where we usually distinguish between partitioning and hierarchical approaches~\cite{jain1999data}. In this work we are mainly concerned with partitioning methods, i.e.\ methods that partition the data into different disjoint sub-sets without any hierarchical structure.
Although clustering methods can be applied to many varying data modalities like user profiles or documents, in this work we consider the specific case of spatial clustering~\cite{grubesic2014spatial}, that normally assumes points to be located in a metric space of dimension $D$, a setting often encountered in practical applications like the facility location problem~\cite{ross1977modeling}.

\subsection{Capacitated Clustering Problems (CCPs)}
Let there be a set of $n$ points $N = \{1, 2, \dots, n\}$ with corresponding feature vectors $x_i \in \IR^D$ of coordinates and a respective weight $q_i \in \mathbb{R}$ associated with each point $i \in N$. 
Further, we assume that we can compute a distance measure $d(x_i, x_j)$ for all pairs of points $i, j \in N$. 
Then we are concerned with finding a set of $K$ capacitated disjoint clusters $c_k \in C, \ c_k \subset N, \ c_k \cap c_l = \emptyset \ \forall k, l \in \{1, \dots, K\}$ with capacities $Q_k > 0$. 
The assignment of points to these clusters is given by the set of binary decision variables $y_{ik}$, which are 1 if point $i$ is a member of cluster $k$ and 0 otherwise.

\subsubsection{Capacitated $p$-Median Problem}
For the CPMP the set of possible cluster \textit{medoids} is given by the set of all data points $N$ and the objective is to minimize the distance (or dissimilarity) $d(x_i,x_{m_k})$ between all $i$ and their cluster \textit{medoid} $m_k$: 
\begin{equation}
	\min\sum_{i \in N}\sum_{k \in K} d(x_i,x_{m_k}) y_{ik}
\end{equation}
s.t.
\begin{align}
	\sum_{k \in K} y_{ik} &= 1, \quad \forall i \in N, \quad \forall k \in K, \\
	\sum_{i \in N} q_i y_{ik} &\leq Q_k, \quad \forall k \in K, \\
	m_k &= \argmin_{m \in c_k} \sum_{i \in c_k} d(x_i,x_{m}), \\
	y_{ik} &\in \{0, 1\}, \quad \forall i \in N, \quad \forall k \in K,
\end{align}
where each point is assigned to only one cluster (2), 
all clusters respect the capacity constraint (3), 
medoids are selected minimizing the dissimilarity of cluster $c_k$ (4)
and $y$ being a binary decision variable (5).

\subsubsection{Capacitated Centered Clustering Problem}
In the CCCP formulation, instead of medoids selected among the data points, \textit{centroids} $\mu_k$ are considered, which correspond to the geometric center of the points assigned to each cluster $c_k$, replacing (4) with
\begin{equation}\label{eq:centers}
	\mu_k = \argmin_{\mu \in \IR^D} \sum_{i \in c_k} d(x_i, \mu),
\end{equation}
which in the case of the Euclidean space for spatial clustering considered in this paper has a closed form formulation: 
\begin{equation}\label{eq:center_update}
	\mu_k = \frac{1}{|c_k|} \sum_{i \in N} x_i y_{ik},
\end{equation}
with $|c_k|$ as cardinality of cluster $c_k$. This leads to the new minimization objective of
\begin{equation}\label{eq:cccp}
	\min\sum_{i \in N}\sum_{k \in K} d(x_i, \mu_k) y_{ik}.
\end{equation}


\section{Related Work}

\paragraph{Clustering Algorithms}
Traditional partitioning methods to solve clustering problems, like the well-known k-means algorithm~\cite{macqueen1967classification} have been researched for more than half a century. Meanwhile, there exists a plethora of different methods including Gaussian Mixture Models~\cite{mclachlan1988mixture}, density based models like DBSCAN~\cite{ester1996density} and graph theoretic approaches~\cite{ozawa1985stratificational}. 
Many of these algorithms have been extended to solve other problem variants like the CCP. 
In particular~\cite{mulvey1984solving} introduce a k-medoids algorithm utilizing a regret heuristic for the assignment step combined with additional re-locations during an integrated local search procedure while~\cite{geetha2009improved} propose an adapted version of k-means which instead uses a priority heuristic to assign points to capacitated clusters. 

\paragraph{Meta-Heuristics}
Apart from direct clustering approaches there are also methods from the operations research community which tackle CCPs or similar formulations like the facility location problem. Different algorithms were proposed modeling and solving the CCP as General Assignment Problem (GAP)~\cite{ross1977modeling}, via simulated annealing and tabu search~\cite{osman1994capacitated}, with genetic algorithms~\cite{lorena2001constructive} or using a scatter search heuristic~\cite{scheuerer2006scatter}.

\paragraph{Math-Heuristics}
In contrast to meta-heuristics, \textit{math-heuristics} combine heuristic methods with powerful mathematical programming solvers like Gurobi~\cite{gurobi}, which are able to solve small scale instance to optimality and have shown superior performance to traditional meta-heuristics in recent studies.
\cite{stefanello2015matheuristics} combine the mathematical solution of the CPMP with a heuristic post-optimization routine in case no optimality was achieved.
The math-heuristic proposed in~\cite{gnagi2021matheuristic} comprises two phases: First, a global optimization phase is executed. This phase alternates between an assignment step, which solves a special case of the GAP as a binary linear program (BLP) for fixed medoids, and a median update step, selecting new medoids $m_k$ minimizing the total distance to all cluster members under the current assignment. This is followed by a local optimization phase  relocating points by solving a second BLP for the sub-set of clusters which comprises the largest unused capacity.
Finally, the PACK algorithm introduced in~\cite{lahderanta2021edge} employs a block coordinate descent similar to the method of~\cite{gnagi2021matheuristic} where the assignment step is solved with Gurobi and the step updating the centers is computed following eq. \ref{eq:center_update} according to the current assignment. 

\paragraph{Deep Clustering}
Since the dawn of deep learning, an increasing number of approaches in related fields is employing deep neural networks. Most approaches in the clustering area are mainly concerned with learning better representations for downstream clustering algorithms, e.g.\ by employing auto-encoders to different data modalities~\cite{tian2014learning,xie2016unsupervised,guo2017deep,yang2017towards}, often trained with enhanced objective functions, which, apart from the representation loss, also include a component approximating the clustering objective and additional regularization to prevent the embedding space from collapsing.
A comprehensive survey on the latest methods is given in~\cite{ren2022deep}.
More recently, deep approaches for constrained clustering have been proposed: \cite{genevay2019differentiable} reformulate the clustering problem in terms of optimal transport to enforce constraints on the size of the clusters. \cite{zhang2021framework} present a framework describing different loss components to include pairwise, triplet, cardinality and instance level constraints into auto-encoder based deep embedded clustering approaches. Finally,  \cite{manduchi2021deep} propose a new deep conditional Gaussian Mixture Model (GMM), which can include pairwise and instance level constraints. 
Usually, the described deep approaches are evaluated on very large, high dimensional datasets like MNIST~\cite{lecun-mnisthandwrittendigit-2010} or Reuters~\cite{xie2016unsupervised}, on which classical algorithms are not competitive. This is in strong contrast to spatial clustering with additional capacity constraints, for which we propose the first deep learning based method.


\section{Proposed Method}
\subsection{Capacitated k-means}\label{cap-k-means}
The capacitated k-means method proposed by~\cite{geetha2009improved} changes the assignment step in Lloyd's algorithm~\cite{lloyd1982least}, which is usually used to compute k-means clustering. 
To adapt the procedure to the CCP, the authors first select the $K$ points with the highest weights $q$ as initial centers, instead of selecting them randomly. Moreover, they introduce priorities $\omega_{ik}$ for each point $i$ by dividing its weight $q_i$ by its distance to the center of cluster $k$:
\begin{equation}\label{eq:heuristic_prio}
	\omega_{ik} = \frac{q_i}{d(x_i, \mu_k)}.
\end{equation}
Then the list of priorities is sorted and nodes are sequentially assigned to the centers according to their priority. 
The idea is to first assign points with large weights to close centers and only then points with smaller weight, which can be more easily assigned to other clusters. 
Next, the centroids are recomputed via the arithmetic mean of the group members (eq.~\ref{eq:center_update}).
The corresponding pseudo code is given in Alg.~\ref{alg:cap-k-means}.

\SetKwComment{Comment}{\cmtsize{//}~}{ }
\SetKwFunction{init}{$\text{init}_{\text{topk\_weights}}$}
\SetKwFunction{allz}{allzero}
\SetKwFunction{rep}{repeat}
\SetKwInOut{Input}{input}
\SetKwInOut{Output}{output}
\DontPrintSemicolon
\SetAlFnt{\small}
\SetAlCapFnt{\small}
\SetAlCapNameFnt{\normalsize}
\begin{algorithm}[tb]
	\caption{Capacitated k-means}\label{alg:cap-k-means}
	\Input{
		$K$, $n$, coordinates $x$, weights $q$, cluster capacity $Q$, convergence condition $\delta$
	}
	\Output{binary assignment matrix Y}
	$M \gets$ \init($x, q, K$) \\				
	\While{not \ $\delta(x, M, Y)$}{	
		$\text{Y} \gets$ \allz($n, K$)			\Comment*[r]{\cmtsize{reset assignment}}
		$\text{Q} \gets$ \rep($Q, K$)		\Comment*[r]{\cmtsize{reset capacities}}
		\ForEach{$i \in N$}{
			compute priorities for all clusters (eq. \ref{eq:heuristic_prio})
		}
		sort priorities, insert into queue $S$\\
		\While{$S$ not empty}{
			get next $i, k$ from $S$\\
			\If{$i$ unassigned  and  $\text{Q}_k \geq q_i$}{
				$\text{Y}_{ik} \gets 1$ 					\Comment*[r]{\cmtsize{cluster assignment}}
				$\text{Q}_k \gets \text{Q}_k - q_i$		\Comment*[r]{\cmtsize{update capacity}}
			}
		}
		\ForEach{$k \in \{1, \dots, K\}$}{
			update centroids via eq. \ref{eq:center_update}
		}
	}
	\textbf{return} Y
\end{algorithm}

\begin{figure*}[tb]
	\centering
	\includegraphics[width=\textwidth]{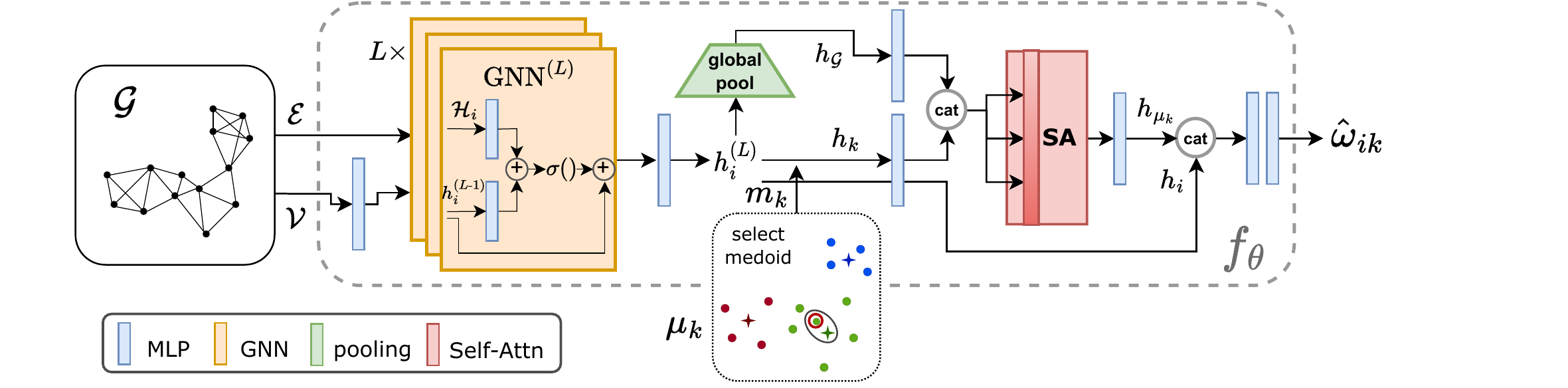}
	\caption{
		Visualization of the neural scoring function architecture. 
	} 
	\label{fig:arch}
\end{figure*}

While this heuristic works, it can easily lead to sub-optimal allocations and situations in which no feasible solution can be found, e.g.\ in cases where many nodes with high weight are located very far from the cluster centers.
To solve these problems we propose several modifications to the algorithm.

\subsection{Neural Scoring Functions}
The first proposed adaption is to learn a neural scoring function $f_{\theta}$ with parameters $\theta$, which predicts the probability of each node $i$ to belong to cluster $k$:
\begin{equation}
	\hat{\omega}_{ik} = f_{\theta}(\G, \mu_k)
\end{equation}
For that purpose we first create a graph representation $\G = (\mathcal{V}, \mathcal{E})$ of the points by connecting each point with its $\mathcal{K}$ nearest neighbors, producing edges $e_{ij} \in \mathcal{E}$ with edge weights $d(x_i, x_j)$. Nodes $v_i \in \mathcal{V}$ are created by concatenating the respective coordinates and weights $[x_i; q_i]$. This graph allows us to define a structure on which the relative spatial information of the different points can be efficiently propagated. We encode $\G$ with the Graph Neural Network (GNN) introduced in \cite{morris2019weisfeiler}, which is able to directly work with edge weights by employing the graph operator defined as
\begin{equation}\label{eq:gnn}
\resizebox{.95\linewidth}{!}{$
	h^{(l)}_i = 
	\sigma\big( \MLP^{(l)}_1(h^{(l-1)}_i)
	+ \MLP^{(l)}_2(\sum_{j\in \mathcal{H}(i)} e_{ji} \cdot h^{(l-1)}_j) \big)
$}
\end{equation}
where $h^{(l-1)}_i \in \IR^{1 \times d_{\text{emb}}}$ represents the embedding of node $i$ at the previous layer $l-1$, $\mathcal{H}(i)$ is the 1-hop graph neighborhood of node $i$, $e_{ji}$ is the directed edge connecting nodes $j$ and $i$, $\MLP_1$ and $\MLP_2$ are Multi-Layer Perceptrons $\MLP: \IR^{d_{\text{emb}}} \to \IR^{d_{\text{emb}}}$ and $\sigma()$ is a suitable activation function. Furthermore, we add residual connections and regularization to each layer. In our case we choose \textit{GELU}~\cite{hendrycks2016gaussian} and \textit{layer normalization}~\cite{ba2016layer} which outperformed \textit{ReLU} and \textit{BatchNorm} in preliminary experiments.
The input layer projects the node features $v_i = [x_i; q_i] \in \IR^{D + 1}$ to the embedding dimension $d_{\text{emb}}$ using a feed forward layer, which then is followed by $L$ GNN layers of the form given in eq. \ref{eq:gnn}.
In order to create embeddings $h_{\mu_k}$ for the centers $\mu_k \in M$ we find the node $j$ closest to $\mu_k$ (corresponding to the cluster medoid $m_k$) and select its embedding $h^{(L)}_j$ as $h_k$. This embedding is concatenated with a globally pooled graph embedding $h_{\G} \in \IR^{d_{\text{emb}}}$:
\begin{equation}
	h_{\G} = \MLP_{\G}\left(\big[\MAX(h^{(L)}); \MEAN(h^{(L)})\big]\right)
\end{equation}
with $\MLP_{\G}: \IR^{2 d_{\text{emb}}} \to \IR^{d_{\text{emb}}}$. 
Next, in order to model the interdependence of different centers, their resulting vectors are fed into a self attention layer (SA)~\cite{vaswani2017attention} followed by another $\MLP_{\mu}: \IR^{2 d_{\text{emb}}} \to \IR^{d_{\text{emb}}}$:
\begin{equation}\label{eq:center_emb}
h_{\mu_k} = \MLP_{\mu}\big( \SA\left(\left[h_{\G}; h_k\right]\right)\big).
\end{equation}
Since we feed a batch of all points $N$ and all currently available centers $\mu$ to our model, they are encoded as a batch of context vectors $h_{\mu_k}$, one for each cluster $k\in \{1, \dots, K\}$. The SA module then models the importance of all cluster center configurations $h_{\mu}$ for each cluster $k$ which allows the model to effectively decide about the cluster assignment probability of every node $i$ to each cluster $c_k$ conditionally on the full information encoded in the latent context embeddings $h_{\mu}$.
Finally, we do conditional decoding 
by concatenating every center embedding with each node and applying a final stack of (element-wise) MLPs.
The architecture of our neural scoring function is shown in Figure \ref{fig:arch}.

\paragraph{Training the model}
We create the required training data and labels by running the math-heuristic solver of~\cite{gnagi2021matheuristic} on some generated datasets to create a good (although not necessarily optimal) partitioning. 
Then we do supervised training using binary cross entropy\footnote{
BCE: $\mathcal{L}(\hat{y}, y) = y \cdot \log \hat{y} + (1-y) \cdot \log(1-\hat{y})$ 
} (BCE) with pairwise prediction of the assignment of nodes $i$ to clusters $k$.

\subsection{Neural Capacitated Clustering}
To fully leverage our score estimator we propose several adaptions and improvements to the original capacitated k-means algorithm (Alg.~\ref{alg:cap-k-means}).
The new method, which we dub \textit{Neural Capacitated Clustering} (NCC) is described in Alg.~\ref{alg:ncc}.

\subsubsection{Order of assignment}
Instead of sorting all center-node pairs by their priority and then sequentially assigning them according to that list, we fix an order given by permutation $\pi$ for the centers and cycle through each of them, assigning one node at a time. 
Since the output of $f_{\theta}$ is the log probability of point $i$ belonging to cluster $k$ and its magnitude does not directly inform the \textit{order} of assignments of different nodes $i$ and $j$ in the iterative cluster procedure, we found it helpful to scale the output of $f_{\theta}$ by the heuristic weights introduced in eq. \ref{eq:heuristic_prio}. Thus, we assign that node $i$ to cluster $k$, which has the highest scaled conditional priority and still can be accommodated considering the remaining capacity $Q_k$.
In case there remain any unassigned points $j$ at the end of an iteration, which cannot be assigned to any cluster since $q_j > Q_k \ \forall k \in K$, we assign them to a dummy cluster $K+1$ located at the origin of the coordinate system. 
We observe in our experiments that already after a small number of iterations usually no nodes are assigned to the dummy cluster anymore, meaning a feasible allocation has been established.
Moreover, since the neighborhood graph $\G$ does not change between iterations, we can speed up the calculation of priorities by pre-computing and buffering the node embeddings $h_i$ and graph embedding $h_{\G}$ in the first iteration.

\subsubsection{Re-prioritization of last assignments}\label{sss:last_assign}
This is motivated by the observation that the last few assignments are the most difficult,
since they have to cope with highly constrained center capacities. Thus, relying on the predefined cyclic order of the centers (which until this point has ensured that approx. the same number of nodes was assigned to each cluster) can lead to sub-optimal assignments in case some clusters have many
nodes with very large or very small weights. To circumvent this problem we propose two different assignment strategies:
\begin{enumerate}
	\item \textit{Greedy:} \ 
	We treat the maximum (unscaled) priority over all clusters as an absolute priority $\bar{\omega}_i$ for all remaining unassigned points $i$:
	\begin{equation}\label{eq:abs_prio}
		\bar{\omega}_i = \max_k \ \hat{\omega}_{ik}
	\end{equation} 
	Then the points are ordered by that priority and sequentially assigned to the closest cluster which can still accommodate them.
	\item \textit{Sampling:} \ 
	We normalize the absolute priorities $\bar{\omega}_i$ of all remaining unassigned points $i$ 
	via the softmax\footnote{softmax: $\sigma(x)_i = \frac{e^{x_i}}{\sum_{j=1}^{n} e^{x_j}}$} function and treat them as probabilities according to which they are sequentially sampled and assigned to the closest cluster which can still accommodate them. This procedure can be further improved by sampling several assignment rollouts and selecting the configuration, which leads to the smallest resulting inertia.
\end{enumerate}
The fraction $\alpha$ of final nodes for which the re-prioritization is applied we treat as a hyperparameter.

\subsubsection{Weight-adapted kmeans++ initialization}
As found in the study of~\cite{celebi2013comparative} standard methods for the selection of seed points for centers during the initialization of k-means algorithms perform quite poorly. This is why the 
k-means++~\cite{arthur2006k} initialization routine was developed, which aims to maximally spread out the cluster centers over the data domain, by sampling a first center uniformly from the data and then sequentially sampling the next center from the remaining data points with a probability equal to the normalized squared distance to the closest already existing center.
We propose a small modification to the k-means++ procedure (called \textit{ckm++}), that includes the weight information into the sampling procedure by simply multiplying the squared distance to the closest existing cluster center by the weight of the data point to sample.

\SetKwComment{Comment}{\cmtsize{//}~}{ }
\SetKwFunction{init}{$\text{init}_{\text{ckm++}}$}
\SetKwFunction{knn}{KNN\_graph}
\SetKwFunction{allz}{allzero}
\SetKwFunction{perm}{random\_perm}
\SetKwFunction{nxt}{get\_next}
\SetKwFunction{rep}{repeat}
\SetKwInOut{Input}{input}
\SetKwInOut{Output}{output}
\DontPrintSemicolon
\SetAlFnt{\small}
\SetAlCapFnt{\small}
\SetAlCapNameFnt{\normalsize}
\begin{algorithm}[tb!]
	\caption{Neural Capacitated Clustering (NCC)}\label{alg:ncc}
	\Input{
		$K$, $n$, coordinates $x$, weights $q$, cluster capacity $Q$, 
		convergence condition $\delta$, scoring function $f_{\theta}$, fraction $\alpha$ 
	}	
	\Output{binary assignment matrix Y}
	$M \gets$ \init($x, q, K$) 				\Comment*[r]{\cmtsize{get seed centers}}
	$\G \gets$ \knn(x)						\Comment*[r]{\cmtsize{create graph}}
	\While{not \ $\delta(x, M, Y)$}{	
		$\text{Y} \gets$ \allz($n, K\text{+}1$)	\\		
		$\text{Q} \gets$ \rep($Q, K$) \\		
		$\pi \gets $ \perm($\{1, \dots, K\}$) \\
		$\hat{\omega} \gets f_{\theta}(\G, M)$  	\Comment*[r]{\cmtsize{compute priorities }}
		sort columns of $\hat{\omega}$ \\
		\While{any $i$ can be assigned}{
			$k \gets \pi$.\nxt() \\
			\ForEach{$i \in N$ sorted by $\hat{\omega}_k$}{
				\If{$i$ unassigned  and  $\text{Q}_k \geq q_i$}{
					$\text{Y}_{ik} \gets 1$ \\ 					
					$\text{Q}_k \gets \text{Q}_k - q_i$	\\		
					\textbf{break foreach}
				}
			}
			\If{fraction of unassigned points $\leq \alpha$}{
				compute absolute priorities $\bar{\omega}$ (eq. \ref{eq:abs_prio})\\
				assign greedily or with sampling (see \ref{sss:last_assign})\\
				\textbf{break while}
			}
		}
		assign any remaining nodes to dummy cluster \\
		\ForEach{$k \in \{1, \dots, K\text{+}1\}$}{
			update centroids via eq. \ref{eq:center_update}
		}
	}
	\textbf{return} Y
\end{algorithm}

\section{Experiments}
We implement our model and the simple baselines in PyTorch~\cite{paszke2019pytorch} version 1.11 and use Gurobi version 9.1.2 for all methods that require it. All experiments are run on a i7-7700K CPU (4.20GHz).
We use $L=4$ GNN layers, an embedding dimension of $d_{\text{emb}}=256$ and a dimension of $d_{\text{h}}=256$ for all hidden layers. 
More details on training our neural scoring function we report in the supplementary.\footnote{
	We open source our code at \url{https://github.com/jokofa/NCC}
}

\subsection{Capacitated Clustering}
\begin{figure}%
	\centering
	\subfloat{{\includegraphics[width=0.23\textwidth]{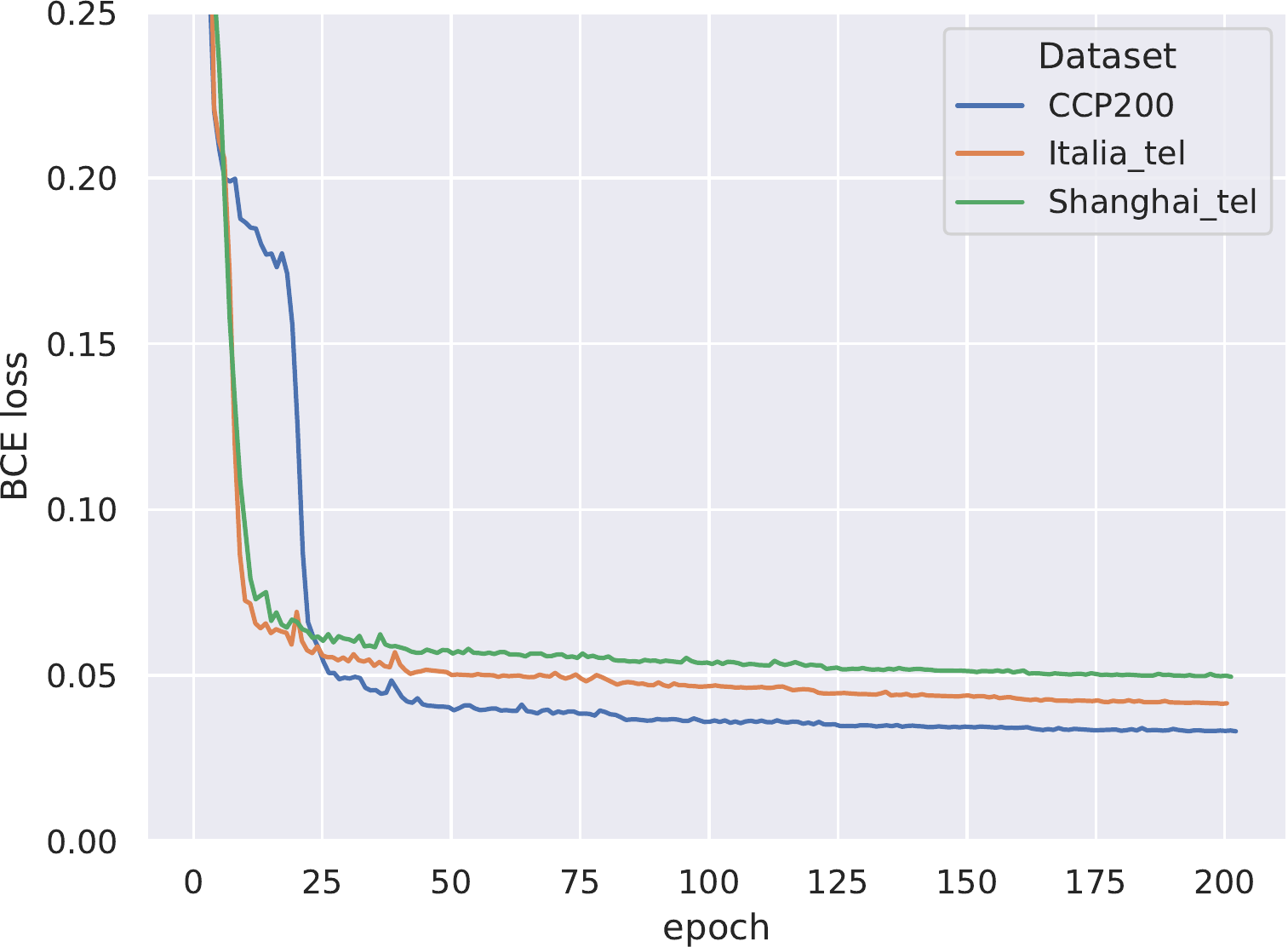} }}%
	\
	\subfloat{{\includegraphics[width=0.23\textwidth]{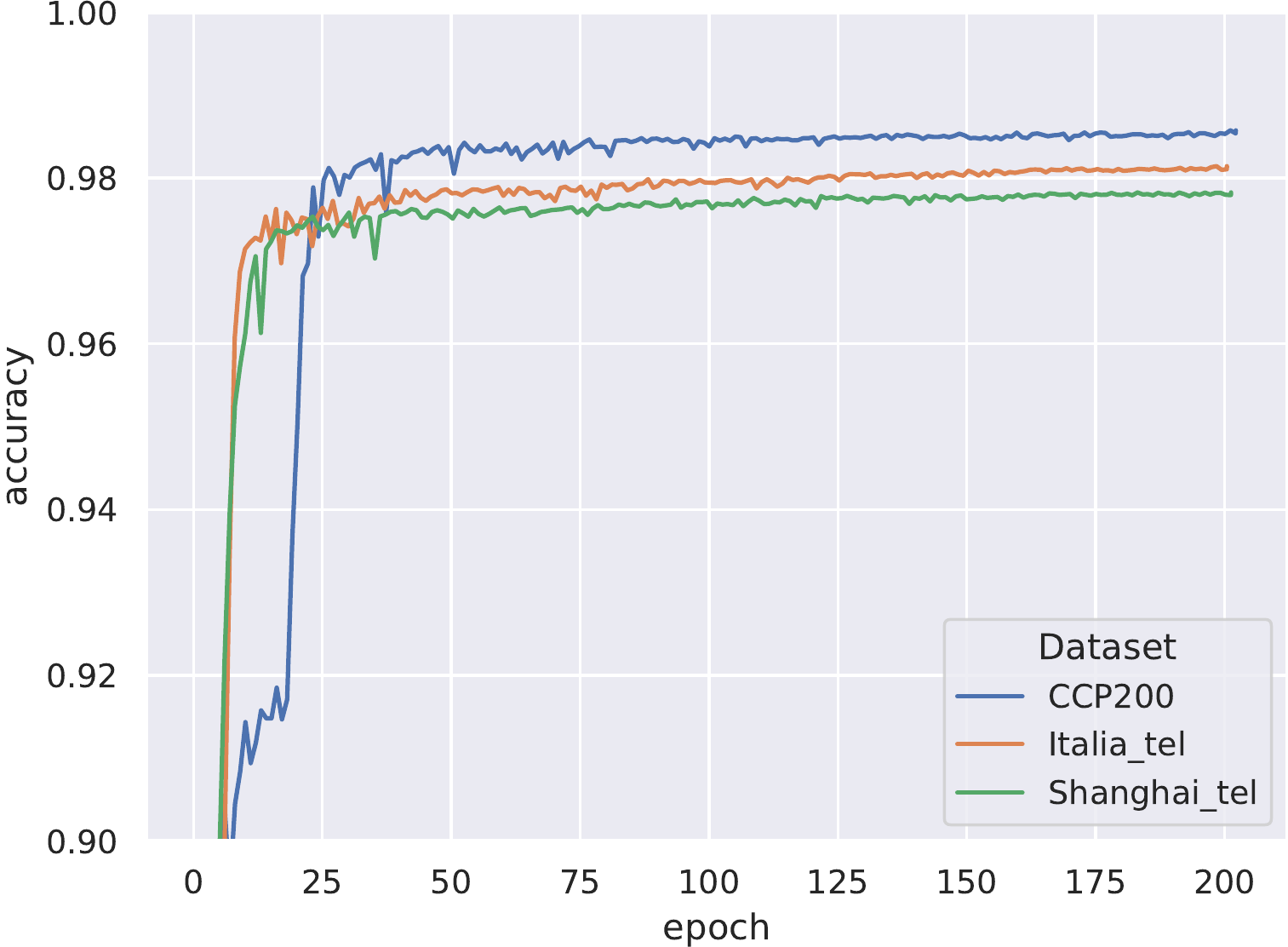} }}%
	\caption{Learning curves and validation accuracy for $f_{\theta}$.}%
	\label{fig:training_loss_acc}%
\end{figure}

For the experiments we use the CCCP formulation of the CCP (eq. \ref{eq:cccp}) which considers centroids instead of medians.
While there are several possible ways to select a useful number $K$ of clusters, like the \textit{Elbow method}~\cite{yuan2019research}, here we adopt a practical approach consisting of solving the problem with the \textit{random} assignment baseline method for a number of seeds and choosing the minimal resulting number of clusters as $K$. 
For the $n=200$ datasets we run all methods for 3 different seeds and report the mean cost with standard deviation and average run times. 
Since Gurobi requires a run time to be specified because it otherwise can take arbitrarily long for the computation to complete, we set reasonable total run times of 3min for $n=200$ and 15min for the original sizes. If Gurobi times out, we return the last feasible assignment if available. Otherwise, we report the result for that instance as infeasible and set its cost to the average cost of the \textit{rnd-NN} baseline. 
Training loss and validation accuracy of our neural scoring function on the different training sets are shown in Figure \ref{fig:training_loss_acc}.
We evaluate our method in a \textit{greedy} and a \textit{sampling} configuration, which we tune on a separate validation set: (g-20-1) stands for 1 greedy rollout for a fraction of $\alpha = 0.2$ and (s-25-32) for 32 samples for $\alpha = 0.25$.

\paragraph{Datasets}
We perform experiments on artificial data and two real world datasets. The artificial data with instances of size $n=200$ we generate based on a GMM. As real world datasets for capacitated spatial clustering we select the well-known \textit{Shanghai Telecom} (ST) dataset~\cite{wang2019edge} which contains the locations and user sessions for base stations in the Shanghai region. In order to use it for our CCP task we aggregate the user session lengths per base station as its corresponding weight and remove base stations with only one user or less than 5min of usage in an interval of 15 days as well as outliers far from the city center, leading to a remaining number of $n=2372$ stations. We set the required number of centers to $K=40$.
The second dataset we assemble by matching the internet access sessions in the call record data of the \textit{Telecom Italia Milan} (TIM) dataset~\cite{barlacchi2015multi} with the Milan cell-tower grid retrieved from OpenCelliD~\cite{opencellid}. After pre-processing it contains $n=2020$ points to be assigned to $K=25$ centers.
We normalize all weights according to $K$ with a maximum capacity normalization factor of 1.1. 
The experiments on the real world data are performed in two different settings: The first setting simply executes all methods on the full dataset, while the second setting sub-samples the data in a random local grid to produce 100 test instances of size $n=200$, with weights multiplied by a factor drawn uniformly from the interval [1.5, 4.0) for ST and [2.0, 5.0) for TIM to produce more variation in the required $K$. The exact pre-processing steps and sub-sampling procedure are described in the supplementary.

\paragraph{Baseline Methods}
\begin{itemize}
	\item \textit{random:} sequentially assigns random labels to points while respecting cluster capacities.
	\item \textit{rnd-NN:} selects $K$ random points as cluster centers and sequentially assigns nearest neighbors to these clusters, i.e.\ points ordered by increasing distance from the center, until no capacity is left.
	\item \textit{topk-NN:} similar to random-NN, but instead selects the $K$ points with the largest weight as cluster centers.
	\item \textit{CapKMeans:} the capacitated k-means algorithm of \cite{geetha2009improved} with \textit{ckm++} initialization which outperformed the original \textit{topk\_weights} initialization.
	\item \textit{PACK:} the block coordinate descent math-heuristic introduced in~\cite{lahderanta2021edge} (using Gurobi).
	\item \textit{GB21:} the two phase math-heuristic proposed by~\cite{gnagi2021matheuristic} also using the Gurobi solver.	
\end{itemize}

\paragraph{Results}
\begin{table}[tb]
	\centering
	\small
	\begin{tabular}{lrrr}
		\toprule
		\textbf{Method}  	& \textbf{Inertia} (\textit{$\pm$}) & \textbf{Time} (\textit{s}) & \textbf{inf.} \%  \\
		\midrule
		random				&	14.35	(3.77) &	0.01	&	0.0	\\
		rnd-NN				&	7.67	(2.35) &	0.01	&	0.0	\\
		topk-NN				&	7.38	(0.00) &	0.01	&	0.0	\\
		GB21				&	0.98	(0.18) &	4.54	&	1.0	\\
		PACK				&	\underline{0.94}	(0.14) &	14.77	&	1.0	\\
		CapKMeans			&	1.30	(0.86) &	2.19	&	5.0	\\
		\midrule
		NCC (g-20-1)		&	0.93	(0.01) &	1.59	&	0.0	\\
		NCC (s-20-64)		&	\textbf{0.92}	(0.01) &	1.83	&	0.0	\\
		\bottomrule
	\end{tabular}
	\caption{Results on generated CCP dataset (100 instances, $n$=200). Best result in \textbf{bold}, second best \underline{underlined}.}
	\label{tab:ccp}
\end{table}

\begin{table}[tb]
	\centering
	\small
	\begin{tabular}{lrrr}
		\toprule
		\textbf{Method}  	& \textbf{Inertia} (\textit{$\pm$}) & \textbf{Time} (\textit{s}) & \textbf{inf.} \%  \\
		\midrule
		\multicolumn{4}{c}{\textbf{Shanghai Telecom (ST)}} \\
		random			&	2.61	(0.88) &	0.02	&	0.0	\\
		rnd-NN			&	1.53	(0.30) &	0.01	&	2.3	\\
		topk-NN			&	1.71	(0.00) &	0.01	&	4.0	\\
		GB21			&	\textbf{0.46}	(0.11) &	17.49	&	3.3	\\
		PACK			&	0.57	(0.16) &	44.47	&	8.7	\\
		CapKMeans		&	0.70	(0.22) &	4.44	&	7.0	\\
		\midrule
		NCC (g-25-1)	&	0.52	(0.02) &	2.87	&	0.0	\\
		NCC (s-25-32)	&	\underline{0.51}	(0.02) &	3.53	&	0.0	\\
		\midrule
		\multicolumn{4}{c}{\textbf{Telecom Italia Milan (TIM)}} \\
		random			&	3.85	(0.54) &	0.02	&	0.0	\\
		rnd-NN			&	2.00	(0.44) &	0.01	&	1.3	\\
		topk-NN			&	2.12	(0.00) &	0.01	&	0.0	\\
		GB21			&	\textbf{0.58}	(0.10) &	14.25	&	2.3	\\
		PACK			&	\underline{0.61}	(0.16) &	70.09	&	4.0	\\
		CapKMeans		&	0.68	(0.14) &	4.05	&	2.3	\\
		\midrule
		NCC (g-20-1)	&	\underline{0.60}	(0.02) &	2.44	&	0.0	\\
		NCC (s-25-128)	&	\textbf{0.58}	(0.01) &	4.09	&	0.0	\\

		\bottomrule
	\end{tabular}
	\caption{Results on sub-sampled ST and TIM datasets (100 instances, $n$=200). Best result in \textbf{bold}, second best \underline{underlined}.}
	\label{tab:st-tim}
\end{table}

\begin{figure*}[tb]
	\centering
	\includegraphics[width=\textwidth]{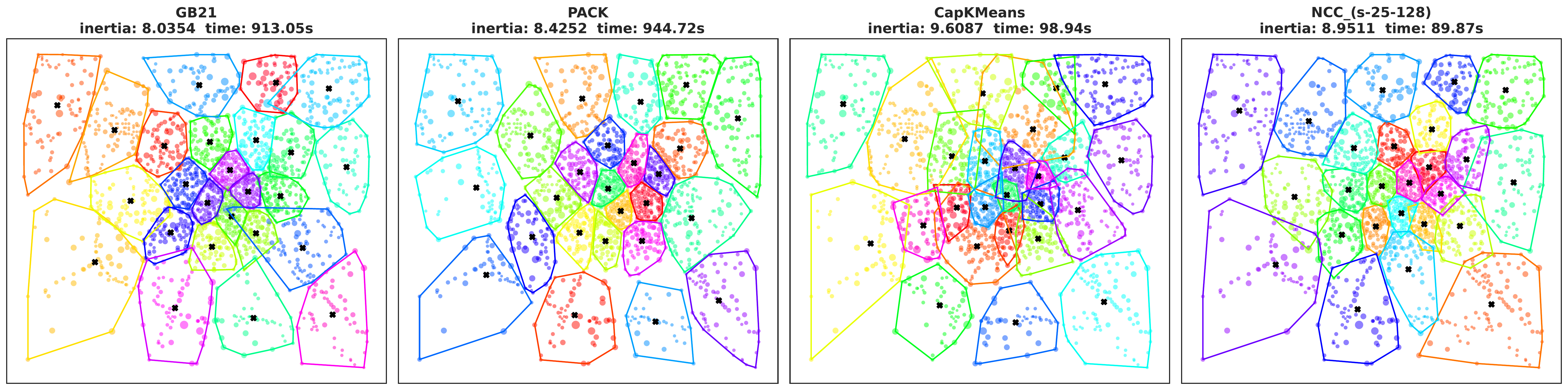}
	\caption{
		Clusters drawn with their convex hulls for the TIM dataset. Black "\textbf{x}" markers represent the cluster centers.
	} 
	\label{fig:TIM}
\end{figure*}
\begin{figure*}[tb]
	\centering
	\includegraphics[width=\textwidth]{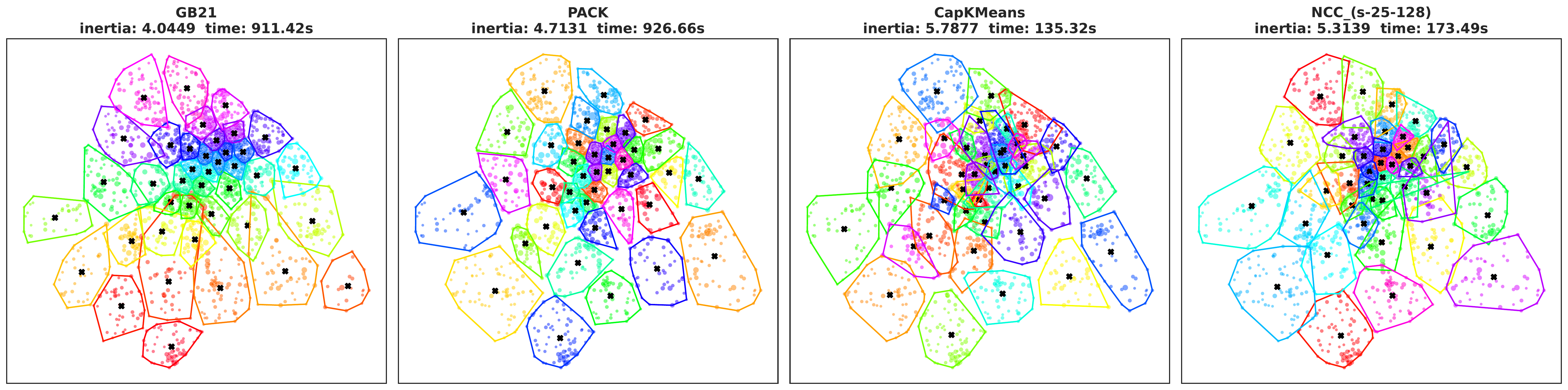}
	\caption{
		Clusters drawn with their convex hulls for the ST dataset. Black "\textbf{x}" markers represent the cluster centers.
	} 
	\label{fig:ST}
\end{figure*}

We evaluate all baselines in terms of \textit{inertia}, which is defined as the total squared distance of all points to their assigned center.
On the generated data (Table \ref{tab:ccp}) our method outperforms all other baselines in terms of inertia while being much faster than all other methods with comparable performance. Furthermore, for three runs with different random seeds our method achieves results with significantly smaller standard deviations and is able to solve all instances within the given time.
Results for the sub-sampled datasets are reported in Table \ref{tab:st-tim}. 
On ST our method beats close to all methods in terms of inertia and is only slightly outperformed by GB21 while being 5$\times$ faster.
On TIM we achieve similar performance compared to GB21 while being much faster and outperform all other approaches. 
In particular, we are more than one order of magnitude faster than the next best performing baseline PACK. Furthermore, our method again achieves very small standard deviation for greedy as well as sampling based assignments, showing that it reliably converges to good (local) optima. 
As expected the random baseline leads to very bad results. The naive baselines \textit{rnd-NN} and \textit{topk-NN} are better but still significantly worse than the more advanced methods, achieving inertia which are 3-7 times higher than that of the best method. 
Compared to \textit{CapKMeans} NCC leads to improvements of 41\%, 37\% and 17\% respectively for the generated, ST and TIM data while achieving even faster run times, showing the impact of our proposed model adaptations.
The inertia and run times for the full datasets are directly reported in the headings of Figures \ref{fig:TIM} and \ref{fig:ST}, which display the cluster assignments for Milan and Shanghai, explicitly drawing the convex hull of each cluster for better visualization. 
While both math-heuristics are able to outperform NCC on these large-scale instances in terms of inertia, our method is close to one order of magnitude faster. Furthermore, the results show that our method is able to find useful cluster structures which especially for TIM are more homogeneous than those of GB21 and show vast improvements compared to \textit{CapKMeans}.

\subsection{Capacitated Vehicle Routing}
To show the efficacy of our approach we extend it to a cluster-first-route-second (C1R2) construction method for Capacitated Vehicle Routing Problems (CVRP). The CVRP is an extension of the traveling salesman problem (TSP) in which $K$ capacitated vehicles have to serve the demand of $n$ customers from a fixed depot node~\cite{toth2014vehicle}. 

\paragraph{Algorithm Modifications}
To adapt our method for the different problem we include an additional MLP in our scoring function, which encodes the depot node, and concatenate the depot embedding with  $h_{\G}$ and $h_k$ in eq. \ref{eq:center_emb}. 
In our algorithm we add the depot node to each cluster $k$ during the center update (Algorithm \ref{alg:ncc}, line 22) and add the distance from each node to the depot to the priority weights. After the clustering we use the fast TSP solver provided by VeRyPy~\cite{rasku2019meta} to route the nodes in each assigned group.

\paragraph{Dataset}
To evaluate our algorithm we choose the benchmark dataset of~\cite{uchoa2017new} which consists of 100 instances of sizes between 100 and 1000 points sampled according to varying distributions (uniform, clustered, etc.) and with different depot positions and weight distributions. We split the benchmark into three sets of problems with size \textbf{N1} ($100 \leq n < 250$), \textbf{N2} ($250 \leq n < 500$) and \textbf{N3} ($500 \leq n$).

\paragraph{Baselines}
In our experiments we compare against several classical C1R2 approaches:
First, the \textit{sweep} algorithm of~\cite{gillett1974heuristic}, which starts a beam at a random point and adds nodes in turn by moving the beam around the depot. We restart the beam at each possible point and run it clock and counter-clock wise. Next, \textit{sweep+}, which instead of routing nodes in the order in which they were passed by the beam, routes them by solving a TSP with Gurobi. The \textit{petal} algorithm introduced in~\cite{foster1976integer} creates "petal" sets by running the sweep algorithm from different starting nodes and then solves a set covering problem with Gurobi to join them.
Finally, for comparison (although not C1R2) the powerful auto-regressive neural construction method \textit{POMO} of~\cite{kwon2020pomo} which is trained with deep reinforcement learning and uses additional instance augmentation techniques. It is evaluated either greedily (g) or with sampling (s) and a beam width of $n$ (size of the instance).

\paragraph{Results}
\begin{table}[tb]
	\centering
	\small
	\begin{tabular}{lrr|rr|rr}
		\toprule
			& \multicolumn{2}{c}{\textbf{N1}} & \multicolumn{2}{c}{\textbf{N2}} & \multicolumn{2}{c}{\textbf{N3}} \\
		\textbf{Method}  	& \textbf{dist} & \textbf{t} (s) & \textbf{dist} & \textbf{t} (s) & \textbf{dist} & \textbf{t} (s) \\
		\midrule
		\textit{sweep}		& 57.2 & 0.65 & 109.7 & 2.21 & 220.7 & 9.80 \\
							& (28.1)	&		& (47.9)		&				& (96.5)		&		\\
		\textit{sweep+}		& 40.8		& 	23.9		& 73.1		& 105.4				& 136.4		& 656.5		\\
							& (28.1)	 	&				& (47.9)		&				& (96.5)		&		\\
		\textit{petal}		& 40.4		& 6.9			& 72.5		& 18.2		& \underline{133.8}	& 86.4		\\
							& (28.1)		&				& (47.9)		&				& (96.5)		&		\\
		\textit{POMO} (g)	& 33.7		& 0.1			& 64.8		& 0.2			& 143.7		& 0.5		\\
							& (24.7)		&				& (44.8)		&				& (87.2)		&		\\
		\textit{POMO} (s)	& \textbf{33.3}		& 1.4		& \textbf{63.8}		& 10.2		& 136.0		& 92.3		\\
							& (24.7)		&				& (44.7)		&				& (87.1)		&		\\
		NCC (g)				& 35.9		& 5.1				& 67.2		& 10.2		& 122.5		& 29.2		\\
							& (\underline{24.0})		&				& (\underline{43.6})		&				& (\underline{84.9})		&		\\
		NCC (s)				& \underline{35.7}		& 7.1				& \underline{66.2}		& 18.5			& \textbf{121.5}	& 39.2	\\
							& (\underline{24.0})		&				& (\underline{43.6})		&				& (\underline{84.9})		&		\\
		\midrule
		$K_{\text{optimal}}$			& \textit{(23.8)}		&		& \textit{(43.5)}		&		& \textit{(84.5)}		&		\\
		\bottomrule
	\end{tabular}
	\caption{Results on the Uchoa benchmark. We report the average total distance, time (sec.) and number of vehicles $K$ (in brackets). $K_{\text{optimal}}$ is the target number of vehicles in the benchmark.}
	\label{tab:uchoa}
\end{table}
As shown in Table \ref{tab:uchoa}, our extended approach performs very competitive on the benchmark, beating all C1R2 approaches from the classical literature and being close to POMO on the small and medium sized instances (N1 and N2) while significantly outperforming it on the large instances (N3). Moreover, our method achieves the smallest fleet size
of all methods, very close to the optimal fleet size $K_{\text{optimal}}$.


\section{Conclusion}
This paper presents the first deep learning based approach for the CCP. 
In experiments on artificial and real world data our method NCC shows competitive performance and fast and robust inference. Moreover, we demonstrate its usefulness as constructive method for the CVRP, achieving promising results on the well-known Uchoa benchmark.

\appendix

\section{Ablation}
\subsection{Algorithm Design}
In this section we present the results of an ablation study to evaluate the usefulness of our adaption to the original capacitated k-means procedure of Geetha et al.~\shortcite{geetha2009improved}.
First, we compare the performance of the original algorithm \textit{CapKMeans} for different initialization routines. The \textit{topk-w} initialization is the one originally proposed by Geetha et al., which simply selects the points with the k largest weights as initial cluster centers. We compare it to the \textit{k-means++}~\cite{arthur2006k} initialization routine, which aims to maximally spread out the cluster centers over the data domain, by sampling a first center uniformly from the data and then sequentially sampling the next center from the remaining data points with a probability equal to the normalized squared distance to the closest already existing center. Moreover, we compare it to our own initialization method \textit{ckm++} that includes the weight information into the sampling procedure of \textit{k-means++} by simply multiplying the squared distance to the closest existing cluster center by the weight of the data point to sample.
Since \textit{topk-w} is deterministic, the seed points for several random restarts are the same and therefore the procedure is only run once. For a fair comparison we also evaluate \textit{k-means++} and \textit{ckm++} for one run 
and report the results in Table \ref{tab:init}.
We sub-sample some $n=200$ instances from the ST dataset and try to solve them in the different setups.
The inertia is measured on the subset of 50 instances for which all five different initializations for \textit{CapKMeans} produced a feasible result. Those results show, that for a single run without restart, the \textit{topk-w} method leads to better performance. However, it is directly clear that several restarts can vastly improve the performance and lead to a significantly reduced number of infeasible instances.

The remainder of the ablation study is concerned with evaluating the impact of the other proposed adaptions which led to our full NCC algorithm. 
The method \textit{CapKMeans-alt} uses the alternative assignment procedure described in the main paper, which cycles through the centers assigning only one point per turn.
Finally, we report the results for the full \textit{greedy} and \textit{sampling} procedures as proposed in the main paper. We evaluate all methods with one run for \textit{topk-w} and eight random restarts for \textit{k-means++} and \textit{ckm++}.

It can be seen that as soon as priorities are computed with the neural scoring function $f_{\theta}$ (for \textit{NCC greedy} and \textit{NCC sampling}), the \textit{ckm++} initialization outperforms the other two methods, while \textit{k-means++} works slightly better in the other two cases. Moreover, we can see that using the neural scoring function significantly improves the algorithm, while simply changing the order of assignment leads to a bit worse results than the original method.

\begin{table}[h!]
	\centering
	\small
	\begin{tabular}{lrrrr}
		\toprule
		\textbf{Init}  	& \textbf{Restarts} & \textbf{Inertia} & \textbf{Avg Run Time} (s) & \textbf{Inf.} \% \\
		\midrule
		\textit{topk-w}		& 1		& 0.825		& 1.88		& 54.0	\\
		\textit{k-means++} 	& 1		& 0.841		& 1.91		& 40.8	\\
		\textit{ckm++} 		& 1		& 0.860		& 1.88		& 44.0	\\
		\textit{k-means++}	& 8		& 0.624		& 2.99		& 2.0	\\
		\textit{ckm++}		& 8		& 0.640		& 3.02		& 3.6	\\
		\bottomrule
	\end{tabular}
	\caption{Ablation on different initialization methods for \textit{CapKMeans}.}
	\label{tab:init}
\end{table}

\begin{figure}[h!]
	\centering
	\includegraphics[width=0.51\textwidth,trim={0.8cm 0.65cm 0.5cm 0.2},clip]{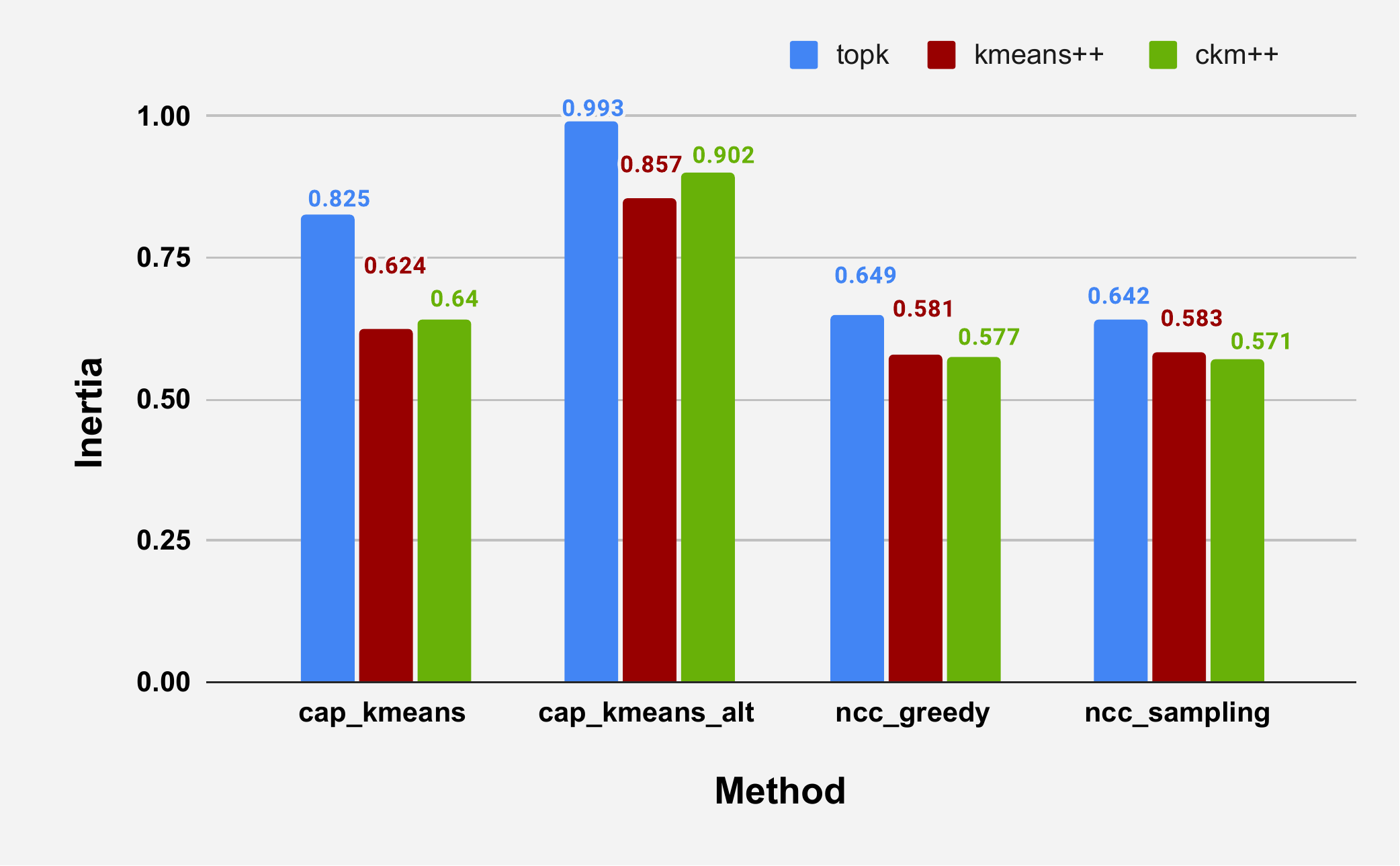}
	\caption{
		Comparison of different initialization methods and algorithm setups in terms of inertia.
	} 
	\label{fig:ablation}
\end{figure}

\begin{figure}[h!]%
	\centering
	\subfloat{{\includegraphics[width=0.33\textwidth]{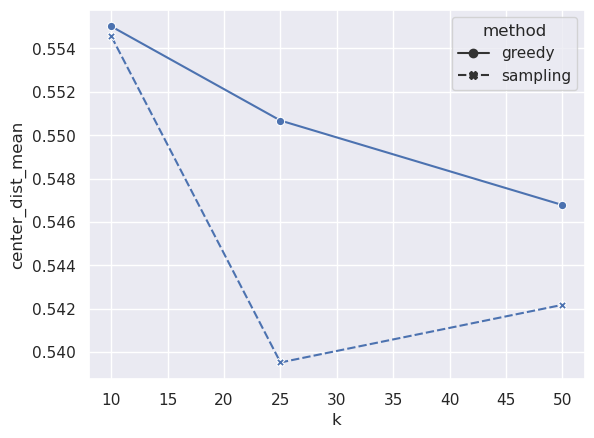} }}%
	\
	\subfloat{{\includegraphics[width=0.33\textwidth]{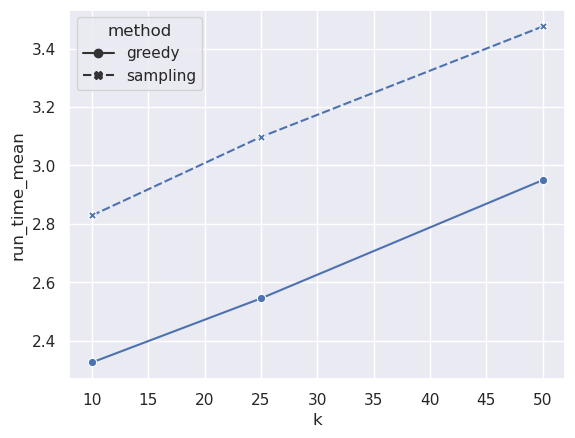} }}%
	\caption{Inertia and run times for different values of k.}%
	\label{fig:knn}%
\end{figure}

We performed a further ablation study regarding the choice of K for the KNN graph. The results are presented in Figure \ref{fig:knn}.
While run time seems to increase very close to linearly with K, larger K does not necessarily increase performance.

\subsection{Generalization to larger and out-of-distribution data}
Furthermore, we performed an additional experiment on a diverse selection of the benchmark instances of \cite{stefanello2015matheuristics} and present the results in table \ref{tab:generalize} below.
Our model trained on the sub-sampled ST data with $n=200$ is able to work well on problem instances from very different distributions and of significantly larger size.
\begin{table}[h!]
	\centering
	\footnotesize
	\begin{tabular}{lrrrrrr}
		\hline
		& \multicolumn{2}{c}{\textbf{NCC (g)}} & \multicolumn{2}{c}{\textbf{NCC (s)}} & \multicolumn{2}{c}{\textbf{GB21}}\\
		\textbf{Instance} 	& \textbf{inert.} & \textbf{t} (\textit{s}) & \textbf{inert.} & \textbf{t} (\textit{s}) & \textbf{inert.} & \textbf{t} (\textit{s}) \\
		\midrule
		ali535\_005		&	2.257	&	4.7		&	2.101	&	7.6		&	\textbf{2.061}	&	5.3		\\
		ali535\_025		&	0.291	&	8.2		&	\textbf{0.284}	&	6.9		&	0.290	&	61.2	\\
		ali535\_050		&	0.166	&	9.3		&	0.158	&	7.4		&	\textbf{0.146}	&	364.6	\\
		fnl4461\_0020	&	9.961	&	62.8	&	\textbf{9.907}	&	58.8	&	10.274	&	241.5	\\
		lin318\_015		&	1.446	&	4.4		&	\textbf{1.424}	&	4.3		&	1.436	&	4.7		\\
		pr2392\_020		&	\textbf{7.307}	&	30.5	&	7.641	&	30.4	&	7.935	&	171.2	\\
		rl1304\_010		&	\textbf{7.762}	&	16.4	&	7.843	&	16.1	&	8.447	&	101.6	\\
		sjc3a\_300\_25	&	0.011	&	2.4		&	0.010	&	3.9		&	\textbf{0.007}	&	16.8	\\
		sjc4a\_402\_30	&	0.014	&	5.3		&	0.014	&	5.4		&	\textbf{0.010}	&	25.3	\\
		spain737\_74	&	0.006	&	2.3		&	0.005	&	10.5	&	\textbf{0.005}	&	238.4	\\
		u724\_010		&	3.891	&	9.1		&	\textbf{3.793}	&	9.0		&	4.148	&	9.5		\\
		u724\_030		&	1.274	&	9.5		&	1.187	&	9.8		&	\textbf{1.174}	&	31.8	\\
		\hline
	\end{tabular}
	\caption{Results on benchmark instances (format: name\_n\_k).
		Our model trained on the sub-sampled ST data with n=200 is able to work well on problem instances from very different distributions and of significantly larger sizes.}
	\label{tab:generalize}
\end{table}

\section{Data}
Apart from the following description of our data related sample and processing steps, we will open source all our data and pre-processing pipelines as jupyter notebooks together with our model code at \url{https://github.com/jokofa/NCC}.

\subsection{Data Generation}
The generated (artificial) data is sampled from a Gaussian Mixture Model where the number $K$ of mixture components is randomly selected between 3 and 12 for each instance. The mean $\mu$ and (diagonal) covariance matrix $\Sigma$ for each component are sampled uniformly from [0, 1]. Weights are also sampled from a standard uniform distribution and re-scaled by a factor of 1.1, i.e.\ for a problem with $K = 3$ components the sum off all weights will be $3/1.1=2.727$. This allows for some flexibility in assigning points to different clusters and is useful to check the ability of the different algorithms to find good assignments in order to minimize the inertia.

\subsection{Real World Data}
\subsubsection{ST}
The \textit{Shanghai Telecom} (ST) dataset~\cite{wang2019edge} contains the locations and user sessions for base stations in the Shanghai region. In order to use it for our CCP task we aggregate the user session lengths per base station as its corresponding weight and remove base stations with only one user or less than 5min of usage in an interval of 15 days. Furthermore, we remove outliers far from the city center, i.e.\ outside of the area between latitude (30.5, 31.75) and longitude (120.75, 122), leading to a remaining number of $n=2372$ stations. We set the required number of centers to $K=40$ and normalize the weights with a capacity factor of 1.1.

\subsubsection{TIM}
The second dataset we assemble by matching the internet access sessions in the call record data of the \textit{Telecom Italia Milan} (TIM) dataset~\cite{barlacchi2015multi} with the Milan cell-tower grid retrieved from OpenCelliD~\cite{opencellid}. To reduce the number of possible cell towers we only select the ones with LTE support.
After pre-processing it contains $n=2020$ points to be assigned to $K=25$ centers.
We normalize all weights according to $K$ with a maximum capacity normalization factor of 1.1.

\subsection{Data Sub-Sampling}
\begin{figure}[tb]
	\centering
	\includegraphics[width=0.49\textwidth]{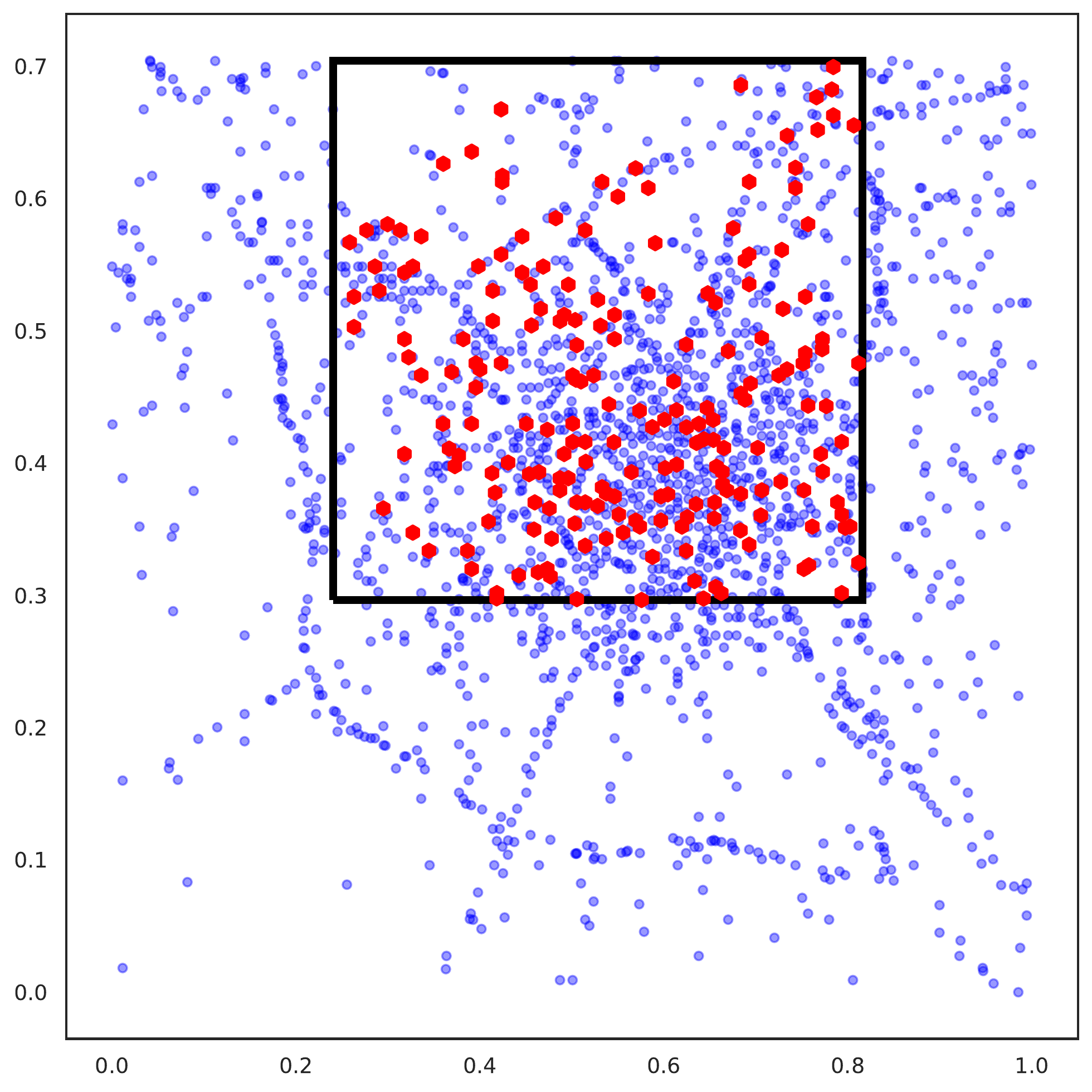}
	\caption{
		Example plot of the used sub-sampling procedure on the TIM dataset. A rectangle (black) of size (0.5*L, 0.5*W) at a random position within the coordinate system is selected. Then $n$ data points (red) are sampled uniformly from within the rectangle. Finally, the weights of the chosen points are re-scaled.
	} 
	\label{fig:subsample}
\end{figure}

In order to replicate the original data distribution of the full real world datasets on a smaller scale, we employ a sub-sampling procedure. Our main concern is that if one would simply sample randomly from the whole grid, then the relative distance between points in the same cluster for small $n$, e.g.\ in our case $n_{\text{sub}} = 200 << n_{\text{full}} = 2020$, would be distorted. Thus, we first select a smaller part of the full point cloud and randomly sub-sample points within that region. To be exact, we first select a random rectangular region of size (0.5*L, 0.5*W), where L and W are the length and width of the full coordinate system in which the original point cloud is contained. Then, we uniformly sample $n_{\text{sub}}$ points from the data points contained in that rectangle. To guarantee sufficient randomness of the sampling process, we require that at least some $n >  n_{\text{sub}}$ points are contained in the rectangular region, from which then $n_{\text{sub}}$ points are selected. Moreover, for $n_{\text{sub}} = 200$ and  $n_{\text{full}} = 2020$ with $K = 25$ clusters as in the TIM dataset, on average there would only be around 2.5 clusters required per sub sample, which does not present a very intersting clustering task. Thus, we rescale the weights of each sub-sample by a factor drawn uniformly from the interval [1.5, 4.0) for ST and [2.0, 5.0) for TIM to produce more variation in the required number of clusters $K$. An exemplary plot of the procedure is given in Figure \ref{fig:subsample}.

\section{Training and Hyperparameters}
Here we report the training regime and hyperparameters used to learn our neural scoring function: \\
Our model is implemented in PyTorch~\cite{paszke2019pytorch} version 1.11.
We use $L=4$ GNN layers, an embedding dimension of $d_{\text{emb}}=256$ and a dimension of $d_{\text{h}}=256$ for all hidden layers of the neural networks. Furthermore, GELU activations~\cite{hendrycks2016gaussian} and layer norm (LN)~\cite{ba2016layer} is employed.
The model is trained for 200 epochs with a mini-batch size of 128 using the Adam optimizer~\cite{kingma2014adam} and a learning rate of $\lambda = 0.001$ which is nearly halved every 40 epochs by multiplying it with a factor of $0.55$. Moreover, we clip the gradient norm above $0.5$. The global seed used for training is 1234. As $\mathcal{K}$ for the KNN graph generation we use $\mathcal{K} = 25$ for the artificial data and the TIM dataset while $\mathcal{K} = 16$ worked better for the ST data.
The corresponding models are trained with training and validation sets with instances of size $n = 200$, either generated artificial data or independently sub-sampled datasets for TIM and ST. The size of the training sets is 4000 for the artificial data and 4900 for both real world datasets. The targets are created by solving the respective instances with the two phase math-heuristic proposed by~\cite{gnagi2021matheuristic} which uses the Gurobi solver~\cite{gurobi}. To solve these datasets in a reasonable time, we set the maximum time for the solver to 32 seconds for the artificial data and 200 seconds for TIM and ST.

For the NCC algorithm we do a small non-exhaustive grid search on the validation data to select the fraction of re-prioritized points $\alpha$ in \{0.05, 0.10, 0.15, 0.20, 0.25\} and the number of samples for the sampling method in \{32, 64, 128\}.
The chosen values are directly reported in the result tables in the main paper.

\section*{Ethical Statement}
There are no ethical issues.

%

\bibliographystyle{named}
\bibliography{references}

\end{document}